\documentclass[letterpaper]{article} 
\usepackage[]{aaai2026}  
\usepackage{times}  
\usepackage{helvet}  
\usepackage{courier}  
\usepackage[hyphens]{url}  
\usepackage{graphicx} 
\usepackage{natbib}  
\usepackage{caption} 
\usepackage[linesnumbered]{algorithm2e}
\usepackage{booktabs}
\usepackage{pgfplots}
\pgfplotsset{compat=newest}
\usepackage{subcaption}
\usepackage{enumitem}
\usepackage{amsmath}
\usepackage{amsfonts}
\usetikzlibrary{patterns}
\usepackage{float}

\newcommand{\spara}[1]{\smallskip\noindent{\bf #1}}

\usepackage{newfloat}
\usepackage{listings}
\DeclareCaptionStyle{ruled}{labelfont=normalfont,labelsep=colon,strut=off} 
\floatstyle{ruled}
\newfloat{listing}{tb}{lst}{}
\floatname{listing}{Listing}
\title{Toward Reasoning on the Boundary: A Mixup-based Approach for \\ Graph Anomaly Detection}
\author{
    Hwan Kim\textsuperscript{\rm 1},
    Junghoon Kim\textsuperscript{\rm 2},
    Sungsu Lim\textsuperscript{\rm 1}\thanks{Corresponding Author.}
}
\affiliations{
    \textsuperscript{\rm 1}Chungnam National University, Daejeon, Republic of Korea \\
    \textsuperscript{\rm 2}UNIST, Ulsan, Republic of Korea \\
}

\begin{document}

\maketitle

\begin{abstract}

While GNN-based detection methods excel at identifying overt outliers, they often struggle with \textit{boundary anomalies}---subtly camouflaged nodes that are difficult to distinguish from normal instances. 
This limitation highlights a fundamental gap in the reasoning capabilities of existing methods.
We attribute this issue to the reliance of standard Graph Contrastive Learning (GCL) on \textit{easy negatives}, which fosters the learning of simplistic decision boundaries. 
To address this issue, we propose {A}{\scriptsize NO}{M}{\scriptsize IX}, a framework that synthesizes informative \textit{hard negatives} by linearly interpolating representations of normal and abnormal subgraphs. 
This graph mixup strategy intentionally populates the decision boundary with hard-to-detect samples. 
Through targeted experimental analysis, we demonstrate that {A}{\scriptsize NO}{M}{\scriptsize IX} successfully separates these boundary anomalies where state-of-the-art baselines fail, as shown by a clear distinction in the score distributions for these challenging cases. 
These findings suggest that synthesizing hard negatives via mixup is a potent strategy for refining GNN representation space, which in turn enhances its reasoning capacity for more robust and reliable graph anomaly detection.
Code is available at \url{https://github.com/missinghwan/ANOMIX}.

\end{abstract}


\section{Introduction}

Graph Neural Networks (GNNs) have emerged as a dominant paradigm for graph anomaly detection (GAD). These approaches, often based on reconstruction objectives~\cite{ding2019deep,kim2022graph} or contrastive learning frameworks~\cite{liu2021anomaly,jin2021anemone}, have significantly advanced the field.
Despite their success, current GNN-based detectors exhibit limitations in identifying \textit{boundary anomalies}---subtly camouflaged nodes that lie in the ambiguous region of the decision boundary between normal and anomalous classes~\cite{kim2023label}.

This shortcoming points to a fundamental gap in the reasoning capabilities of these models.
For instance, contrastive methods that rely on simple augmentations often fail to assign sufficiently distinct scores to anomalies that share significant local structures with normal instances~\cite{jin2021anemone,duan2023graph}. Similarly, reconstruction-based approaches can proficiently rebuild such nodes due to their high degree of structural normalcy, inadvertently masking their subtle attribute-level deviations~\cite{luo2022comga}. Consequently, these models are proficient at spotting overt deviations but often fail to capture the characteristic of nuanced patterns near boundaries.

We attribute this weakness to the training schemes of prevalent GCL methods. 
The core issue lies in their reliance on contrasting an anchor against \textit{easy negatives}, which are typically generated via simple augmentations such as random node or edge perturbations~\cite{zhu2020deep,you2020graph}, processes that encourage learning a simplistic, low-resolution decision boundary. 
Recognizing this limitation, recent works have begun to explore more sophisticated negative sampling, such as mapping latent boundaries or employing adversarial training, yet a principled method for synthesizing informative hard negatives for boundary anomalies remains an open challenge~\cite{lu2025agmixup}.
Consequently, to enhance a model's ability to reason and identify subtle anomalies, it is crucial to train models on more informative \textit{hard negatives} that effectively populate this boundary region~\cite{liu2025dual}.

\begin{figure}[t]
    \centering
    \includegraphics[width=0.85\linewidth]{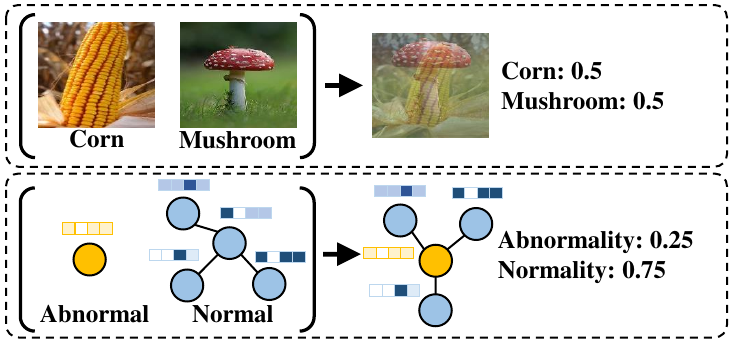}
    \caption{Generating hard negatives for image (top) and graph (bottom).}
    \label{fig:mixup_example}
\end{figure}

Our approach is theoretically motivated by the principle of Vicinal Risk Minimization (VRM)~\cite{chapelle2000vicinal}, which posits that model generalization can be enhanced by training on virtual samples drawn from the vicinity of observed data. 
The Mixup technique~\cite{zhang2018mixup} serves as a simple yet effective implementation of this principle, creating such virtual samples by linearly interpolating pairs of data points.
This principle has been successfully extended to graph-structured data; for instance, interpolating graph representations has been shown to improve performance in node and graph classification~\cite{wang2021mixup,han2022g}.
However, while prior graph mixup techniques have shown promising results in classification tasks, their application to the unique challenges of GAD—specifically for composing informative negative samples—has been largely unexplored.

To bridge this gap, we introduce {A}{\footnotesize NO}{M}{\footnotesize IX}, an approach that synthesizes hard negatives by \textit{mixing} the representations of normal and abnormal subgraphs (see Fig.~\ref{fig:mixup_example}). 
Our graph mixing strategy, {A}{\footnotesize NO}{M}{\footnotesize IX-M}, serves to populate the decision boundary with challenging training instances, thereby compelling the GNN to learn a more refined and robust class separation.
Our primary contributions are:
(1) We propose the first graph mixing strategy tailored for hard negative generation in the context of GAD. 
(2) Through targeted experimental analysis, we demonstrate that our approach significantly bolsters a GNN's ability to identify boundary anomalies, thus enhancing its reasoning capacity for more robust detection.

\section{{A}{\footnotesize NO}{M}{\footnotesize IX}: Hard Negative Generation via Graph Mixup}

\begin{figure}[t]
    \centering
    \includegraphics[width=0.99\linewidth]{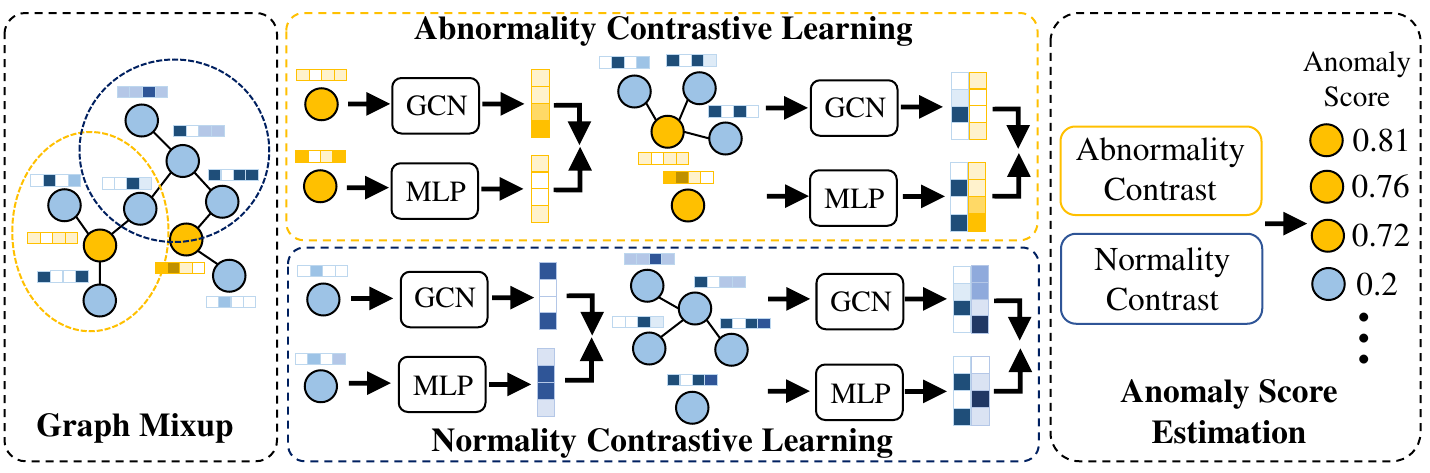}
    \caption{An overview of the {A}{\footnotesize NO}{M}{\footnotesize IX} framework.}
    \label{fig:ANOMIX}
\end{figure}

The {A}{\footnotesize NO}{M}{\footnotesize IX} framework comprises two main components: a graph mixup module for hard negative synthesis, and a multi-level contrastive learning module for representation learning. The overall architecture is depicted in Figure~\ref{fig:ANOMIX}.

\subsection{Graph Mixup ({A}{\footnotesize NO}{M}{\footnotesize IX-M})}

The core of our approach is {A}{\footnotesize NO}{M}{\footnotesize IX-M}, a strategy engineered to synthesize informative hard negatives. For a given target node, {A}{\footnotesize NO}{M}{\footnotesize IX-M} constructs two contextual subgraphs. The normal context, represented by the ego-net $G^{no}$, is sampled via random walks initiated from the target node itself. To construct the abnormal context, $G^{ab}$, we employ a semi-supervised setting by leveraging a minimal set of labeled anomalies. Specifically, $G^{ab}$ is an ego-net sampled from walks originating from an anchor node randomly chosen from this limited set of known anomalies.

Subsequently, {A}{\footnotesize NO}{M}{\footnotesize IX-M} generates a hard negative sample, $G_{mix}$, by linearly interpolating the representations of these two subgraphs:
\begin{equation} \label{eq:mixup}
    G_{mix} = \lambda {G}^{ab} + (1-\lambda){G}^{no},
\end{equation}
where the mixing coefficient $\lambda$ is drawn from a Beta distribution, $\lambda \sim \text{Beta}(\alpha, \alpha)$. 
This choice is conventional in mixup literature, as the Beta distribution naturally constrains the sampled values to [0, 1] range required for interpolation and offers fine-grained control over the distribution of mixing ratio through the hyperparameter $\alpha$.
Additionally, we also employ a feature masking strategy, setting the target node's features as zero within the input subgraphs to prevent information leaking~\cite{jin2021anemone,liu2021anomaly}.

\subsection{Contrastive Learning and Anomaly Scoring}

With the synthesized hard negatives, we train the model using a multi-level contrastive learning objective. 
The framework is designed to learn discriminative representations at both the node-level and subgraph-level. 
At the node-level, the objective is to distinguish the original target node's embedding from the embedding of its masked counterpart within the subgraph context. 
At the subgraph-level, the model contrasts the target node's embedding against a readout summary of the entire subgraph.

To implement this, we employ a GNN encoder to obtain representations for the nodes within a given subgraph. A bilinear scoring function then assesses the similarity between positive pairs (e.g., target node and its context) and negative pairs (e.g., target node and a corrupted context). The model is optimized using a contrastive loss objective, which aims to maximize the scores for positive pairs while minimizing them for negative ones across both normal and mixed (abnormal) views.

For inference, the final anomaly score for a given node is derived from the aggregated outputs of this contrastive framework over multiple stochastic sampling rounds. 
Intuitively, a node is deemed anomalous if it consistently yields a high discrepancy between its positive and negative similarity scores. 
We aggregate these score differences, incorporating both their mean and standard deviation to capture not only the magnitude but also the instability of the scores—a common trait of anomalous nodes.

\begin{table}[t]
\small
\centering
\caption{The statistics of the datasets.}
\label{tab:dataset_statistics}
\resizebox{0.99\columnwidth}{!}{%
\begin{tabular}{@{}lccccc@{}}
\toprule
\phantom{1} Datasets & Nodes & Edges & Attributes & Anomalies & R/I\phantom{1} \\ \midrule
\phantom{1} Cora     & 2,708    & 5,429    & 1,433  & 150  & I  \\
\phantom{1} CiteSeer & 3,327    & 4,732    & 3,703  & 150  & I  \\
\phantom{1} Pubmed   & 19,717   & 44,338   & 500    & 600  & I  \\
\phantom{1} ACM      & 16,484   & 164,630  & 8,337  & 600  & I  \\ 
\phantom{1} Facebook & 4,039    & 88,234   & 576    & 27   & R  \\ 
\phantom{1} Amazon   & 10,244   & 175,608  & 25     & 693  & R  \\ \bottomrule
\end{tabular}
}
\end{table}

\begin{table}[t]
\small
\centering
\caption{Comparison of {A}{\footnotesize NO}{M}{\footnotesize IX} and baselines (AUC). The highest AUC value is boldfaced and second highest AUC value is underlined.}
\label{tab:auc_results}
\resizebox{0.99\columnwidth}{!}{%
\begin{tabular}{@{}lcccccc@{}}
\toprule
 Methods  & Cora & \begin{tabular}[c]{@{}c@{}}Cite\\Seer\end{tabular} & Pubmed & ACM & \begin{tabular}[c]{@{}c@{}}Face\\ book\end{tabular} & Amazon   \\ \midrule
 DOMINANT & 81.55 & 82.51 & 80.81 & 76.01 & 56.97 & 59.38  \\
 CoLA & 87.78 & 89.68 & \underline{95.12} & 82.37 & \underline{83.44} & 58.42   \\
 ANEMONE & 90.57 & \underline{91.89} & 94.48 & \underline{86.90} & 76.93 & 58.59 \\
 ComGA & 88.40 & 91.67 & 92.20 & 84.96 & 61.67 & 57.98   \\
 LHML & 84.40 & 80.20 & 83.50 & 82.10 & 78.26 & \underline{61.37}  \\
 GRADATE & \underline{92.37} & 83.13 & 82.21 & 84.70 & 64.92 & 51.29  \\ \midrule
 DeepSAD & 56.70 & 54.60 & 51.10 & 53.00 & 56.72 & 51.29   \\
 GDN & 83.17 & 85.48 & 84.72 & 78.12 & 76.26 & 49.48   \\
 Semi-GNN & 67.65 & 81.74 & 66.30 & 84.90 & 52.88 & 45.85   \\ 
 MetaGAD & 83.91 & 86.28 & 83.47 & 80.82 & 78.38 & 52.47 \\ \midrule
 \textbf{{A}{\footnotesize NO}{M}{\footnotesize IX}} (ours) & \textbf{93.27} & \textbf{94.14} & \textbf{95.58} & \textbf{91.04} & \textbf{91.88} & \textbf{66.86}  \\ \bottomrule
\end{tabular}
}
\end{table}

\begin{figure*}[t]
    \centering
    \includegraphics[width=0.99\linewidth]{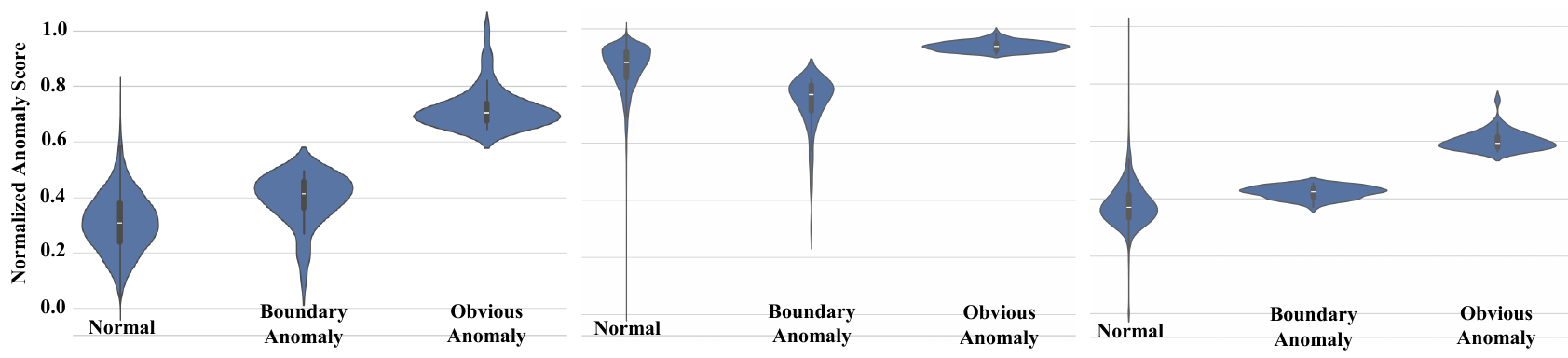}
    \includegraphics[width=0.99\linewidth]{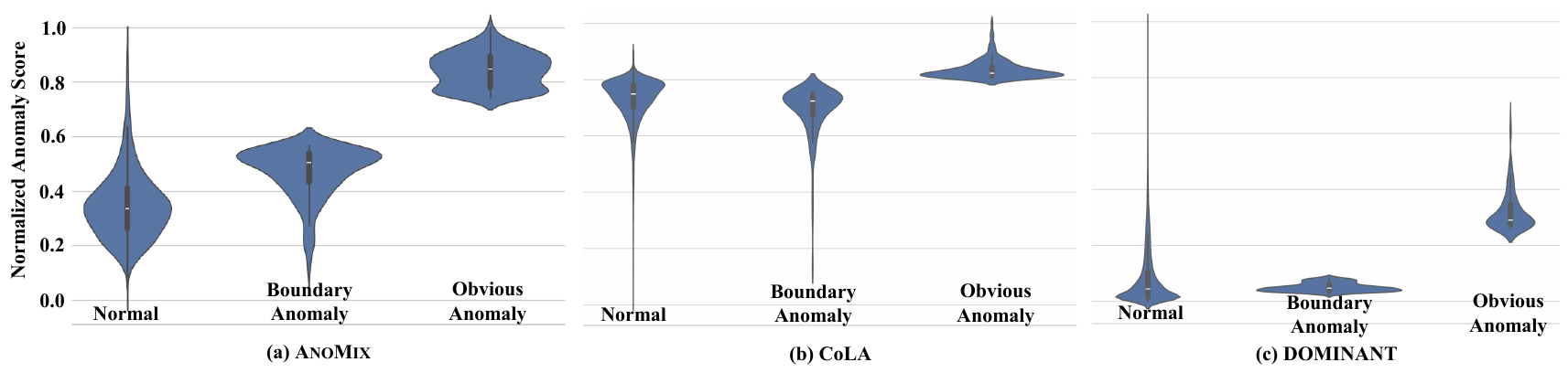}
    \caption{Anomaly score distributions on Citeseer (up) and Pubmed (down). The baseline model (b) and (c) fail to distinguish Boundary Anomalies from Normal nodes. In contrast, {A}{\footnotesize NO}{M}{\footnotesize IX} (a) successfully identifies these nodes by assigning significantly higher scores, demonstrating its robustness in detecting hard-to-find anomalies.}
    \label{fig:violin_plot}
\end{figure*}

\section{Experiments}

We evaluate {A}{\footnotesize NO}{M}{\footnotesize IX} on six benchmark datasets and compare it against 10 state-of-the-art methods.

\spara{Datasets.}
We choose six widely-used real world networks ranging from citation and social networks to online commercial network, including ACM, Cora, CiteSeer, Pubmed, Amazon~\cite{dou2020enhancing}, and Facebook~\cite{xu2022contrastive}. 
In addition, since there are no anomalies in the citation networks (first four datasets), we conduct the same anomaly injection process as previous works~\cite{ding2019deep,liu2021anomaly,jin2021anemone,luo2022comga}.
The other two datasets contain real anomalies.
The statistics are summarized in Table~\ref{tab:dataset_statistics}.
The column R/I is type of anomaly, either `Injected' or `Real'.

\spara{Baselines.}
We compare {A}{\footnotesize NO}{M}{\footnotesize IX} against 10 SOTA baselines, including reconstruction-based (e.g., DOMINANT \cite{ding2019deep}, ComGA \cite{luo2022comga}, LHML \cite{guo2022learning}), GCL-based (e.g., CoLA \cite{liu2021anomaly}, ANEMONE \cite{jin2021anemone}, GRADATE \cite{duan2023graph}), and semi-supervised methods (e.g., DeepSAD~\cite{ruffdeep}, GDN~\cite{ding2021few}, Semi-GNN~\cite{kumagai2021semi}, MetaGAD~\cite{xu2023metagad}).


\subsection{Overall Performance}
Overall, {A}{\footnotesize NO}{M}{\footnotesize IX} outperforms all state-of-the-art baselines across all datasets, achieving up to \textbf{8.44}\% higher AUC (Table~\ref{tab:auc_results}). While existing GCL methods struggle in semi-supervised settings, our approach efficiently integrates GCL with hard negatives. On the Pubmed dataset, {A}{\footnotesize NO}{M}{\footnotesize IX} performs better than CoLA but shows a less distinct advantage, more likely due to the synthetic nature of its anomalies. Pubmed is a sparse network (average degree $<$ 3) where anomalies are injected as cliques of $c=15$ fully connected nodes~\cite{skillicorn2007detecting}. In such structure, hidden anomalies that blend with normal nodes are rare, limiting the advantage of our method in detection.

Notably, our results in Table~\ref{tab:auc_results} reveal that leading semi-supervised methods do not consistently outperform unsupervised methods, indicating a risk of overfitting to the limited known anomalies and poor generalization to unseen anomaly cases. In contrast, unsupervised methods learn more generalized representations. 
{A}{\footnotesize NO}{M}{\footnotesize IX} builds upon this strength by not merely relying on a few labeled instances, but by actively synthesizing hard negatives that explore the very boundary between normal and abnormal contexts. This strategy fosters more robust and generalizable decision boundary, which is critical for detecting subtly camouflaged anomalies.

\spara{Performance analysis.} 
The performance improvement of {A}{\footnotesize NO}{M}{\footnotesize IX} on ACM and Facebook datasets stems from its effectiveness against their distinct types of hard-to-detect anomalies.
In the dense ACM network, where injected structural anomalies can be masked by high connectivity, our mixup strategy forces the model to learn more discriminative structural representations by contrasting a node's local context with a known abnormal subgraph. Conversely, on Facebook, which contains real-world anomalies often lacking overt structural irregularities, {A}{\footnotesize NO}{M}{\footnotesize IX} outperforms by synthesizing borderline cases. This process makes the model sensitive to slightly different attribute-based deviations, a scenario in which standard GCL models that primarily rely on structural perturbations tend to perform poorly.

\subsection{Analysis on Boundary Anomalies}
The central thesis of our work is that {A}{\footnotesize NO}{M}{\footnotesize IX} excels in identifying hard-to-detect boundary anomalies. To investigate this, we perform a targeted analysis centered on the score distributions of different models.

\spara{Defining boundary and obvious anomalies.}
To quantitatively analyze performance on challenging cases, we first establish a baseline to partition the anomalous nodes. We train a standard GCL model, CoLA, and categorize the ground-truth anomalies based on its output scores. Anomalies falling in the bottom 30th percentile of scores are labeled as \textit{`Boundary`}, representing the most difficult-to-detect cases. Conversely, those in the top 30th percentile are labeled as \textit{`Obvious`}.

\spara{Results.}
Figure~\ref{fig:violin_plot} presents the core findings of our analysis on the Pubmed dataset. 
As shown in Figure~\ref{fig:violin_plot}(b) and (c), the score distributions of the baseline models for `Boundary` anomalies heavily overlap with that of `Normal` nodes. This confirms that these anomalies are indeed challenging for standard GNN-based methods, as they are frequently assigned low anomaly scores and thus misclassified.

On the other hand, Figure~\ref{fig:violin_plot}(a) illustrates that {A}{\footnotesize NO}{M}{\footnotesize IX} successfully resolves this ambiguity. The score distribution for the `Boundary` group is clearly separated from the `Normal` group and is shifted significantly towards the `Obvious Anomaly` distribution. 
Notably, this trend is not specific to Pubmed in Figure~\ref{fig:violin_plot}(down) as we observe a consistent pattern on the Citeseer in Figure~\ref{fig:violin_plot}(up) as well, which suggests our findings are generalizable and not merely an artifact of a particular dataset's characteristics.
This result provides strong evidence that our mixup-based hard negatives compel the model to learn more refined decision boundary, enhancing its ability to identify ambiguous cases that other baseline models fail to capture.

\spara{Qualitative analysis.}
A case study of a boundary anomaly node from the Cora dataset, which was misclassified by CoLA but correctly identified by {A}{\footnotesize NO}{M}{\footnotesize IX}, provides qualitative insights. The node exhibits subtle irregularities; it is structurally peripheral and its attributes slightly deviate from its normal neighbors. Standard GCL models like CoLA, which rely on simple perturbations for negative sampling, overlook these weak anomalous signals.
In contrast, {A}{\footnotesize NO}{M}{\footnotesize IX} effectively captures these subtle anomalous signals. By creating a synthetic hard negative that interpolates the target node’s local context with that of a known anomaly, our method forces the model to reason about a challenging scenario: a structurally plausible node exhibiting attribute patterns from the known anomalies. 
This process makes the model sensitive to the node's subtle deviations by associating them with proposed abnormal context, leading to its successful detection and demonstrating more precise reasoning capability.

\subsection{Ablation Study on Mixup Strategy} 

To validate that the performance gains of {A}{\footnotesize NO}{M}{\footnotesize IX} are directly attributable to our proposed mixup strategy, we conduct a targeted ablation study. We compare our full model against two variants: (1) \texttt{w/o Mixup}, which reduces to a standard GCL framework, and (2) \texttt{w/ Random Mixup}, a naive approach that interpolates randomly chosen subgraphs instead of targeted normal-abnormal pairs.
As reported in Table~\ref{tab:ablation_results}, the \texttt{w/o Mixup} variant yields the lowest performance across all datasets. This result underscores our central claim that a conventional contrastive objective alone is insufficient for capturing the complex decision boundaries necessary to identify boundary anomalies.
More revealing is the comparison with the \texttt{w/ Random Mixup} variant. While this untargeted mixing provides a marginal improvement over no mixup, our principled strategy of mixing specifically `normal` and `abnormal` contexts yields better results.

This evidence strongly suggests that the key to success of {A}{\footnotesize NO}{M}{\footnotesize IX} is not mixing itself, but rather our \textit{well-designed} strategy for synthesizing hard negatives that explicitly populate the decision boundary. It confirms that this targeted approach is the crucial factor in enhancing the model's ability to detect boundary anomalies.

\begin{table}[t]
\small
\centering
\caption{Effect of different mixup strategies (AUC).} 
\label{tab:ablation_results}
\resizebox{0.99\columnwidth}{!}{%
\begin{tabular}{@{}lcccc@{}}
\toprule
\phantom{1}Methods    & Cora & CiteSeer & Pubmed & ACM\phantom{1} \\ \midrule
\phantom{1}{A}{\scriptsize NO}{M}{\scriptsize IX} w/o Mixup & 89.98 & 92.26 & 90.96 & 82.45\phantom{1} \\
\phantom{1}{A}{\scriptsize NO}{M}{\scriptsize IX} w/ Random Mixup & 92.26 & 93.23 & 94.68 & 87.86\phantom{1} \\ 
\phantom{1}\textbf{{A}{\scriptsize NO}{M}{\scriptsize IX}}       & \textbf{93.27} & \textbf{94.14} & \textbf{95.58} & \textbf{91.04}\phantom{1} \\ \bottomrule
\end{tabular}
}
\end{table}

\section{Discussion and Future Work}

Our work demonstrates that by explicitly training on the decision boundary via mixup, a practical implementation of the VRM principle, GNNs can identify better such ambiguous cases. 
Aligned with the GCLR workshop's theme, a key future direction is adapting the {A}{\footnotesize NO}{M}{\footnotesize IX} framework for more complex structures.
Specifically, investigating how mixup can be defined for heterogeneous graphs, multi-relational graphs, or dynamic graphs could address a wider range of sophisticated graph anomaly detection problems.
Furthermore, current implementation samples the mixing coefficient from a static Beta distribution. A more advanced approach could involve an adaptive strategy where the mixing coefficient $\lambda$ is dynamically adjusted based on the characteristics of the subgraphs being interpolated, potentially creating to even more informative hard negatives.

\section{Conclusion}

In this work, we addressed a critical limitation in GNN-based GAD: the difficulty of identifying boundary anomalies that reside in the ambiguous region between normal and anomalous classes. 
We introduced {A}{\footnotesize NO}{M}{\footnotesize IX}, a novel framework that synthesizes hard negatives through a tailored graph mixup strategy. 
Our experimental analysis demonstrates that {A}{\footnotesize NO}{M}{\footnotesize IX} significantly enhances a GNN's ability to detect these challenging, hard-to-distinguish nodes where standard models typically fail. 
These findings suggest that a well-designed hard negative generation scheme can effectively sharpen the GNN's decision boundary, representing a tangible step toward more robust reasoning capabilities in graph anomaly detection.

\bibliography{aaai2026}


\end{document}